\documentclass{article}

\usepackage[preprint]{neurips_2024}

\usepackage{natbib}
\usepackage[utf8]{inputenc} 
\usepackage[T1]{fontenc}    
\usepackage{hyperref}       
\usepackage{url}            
\usepackage{booktabs}       
\usepackage{amsfonts}       
\usepackage{nicefrac}       
\usepackage{microtype}      
\usepackage{xcolor}         
\usepackage{graphicx}
\graphicspath{ {./images/} }

\title{LLMs achieve adult human performance on higher-order theory of mind tasks}

\author{
Winnie Street$^{1*}$ \hspace{0.1cm} John Oliver Siy$^1$ \hspace{0.1cm} Geoff Keeling$^1$ \hspace{0.1cm} Adrien Baranes$^2$  \\ \hspace{0.1cm} \textbf{Benjamin Barnett}$^1$
\textbf{Michael Mckibben}$^3$ \hspace{0.1cm} \textbf{Tatenda Kanyere}$^4$ \hspace{0.1cm} \\ \textbf{Alison Lentz}$^1$ \hspace{0.1cm} \textbf{Blaise Aguera y Arcas}$^1$ \hspace{0.1cm} \textbf{Robin I. M. Dunbar}$^5$ \\
$^1$Google Research \hspace{0.1cm}
 $^2$ Google DeepMind 
\hspace{0.1cm} $^3$Applied Physics Lab, Johns Hopkins University \\ \quad $^4$Work done at Google Research via Harvey Nash \\ \quad $^5$ Department of Experimental Psychology, University of Oxford \\
\texttt{istreet@google.com}\\
}

\begin{document}

\maketitle

\begin{abstract}
  This paper examines the extent to which large language models (LLMs) have developed higher-order theory of mind (ToM); the human ability to reason about multiple mental and emotional states in a recursive manner (e.g. I \textit{think} that you \textit{believe} that she \textit{knows}). This paper builds on prior work by introducing a handwritten test suite -- Multi-Order Theory of Mind Q\&A -- and using it to compare the performance of five LLMs to a newly gathered adult human benchmark. We find that GPT-4 and Flan-PaLM reach adult-level and near adult-level performance on ToM tasks overall, and that GPT-4 exceeds adult performance on 6th order inferences. Our results suggest that there is an interplay between model size and finetuning for the realisation of ToM abilities, and that the best-performing LLMs have developed a generalised capacity for ToM. Given the role that higher-order ToM plays in a wide range of cooperative and competitive human behaviours, these findings have significant implications for user-facing LLM applications.
\end{abstract}

\section{Introduction}

Theory of Mind (ToM) is the ability to infer and reason about the mental states of oneself and others \citep{premack1978does, wimmer1983beliefs, wellman2001meta}. ToM is central to human social intelligence: it enables humans to predict and influence behaviour \citep{humphrey1976social, wellman1988young, hooker2008mentalizing}. 

Large Language Models (LLMs) exhibit some ToM competency \citep{kosinski2023theory, bubeck2023sparks, shapira2023clever}. Most of the literature on LLM ToM has focused on 2nd-order ToM \citep{sap2022neural, kosinski2023theory, gandhi2024understanding, shapira2023clever}, where the ‘order of intentionality’ (hereafter, `order') is the number of mental states involved in a ToM reasoning process (i.e. a third-order statement is “I \textit{think} you \textit{believe} that she \textit{knows}”). Yet LLMs are increasingly leveraged for multi-party social interaction contexts which require LLMs to engage in \textit{higher order} ToM reasoning \citep{wang2023survey, park2023generative}.

In this paper, we examine LLM ToM from orders 2-6. We introduce a novel benchmark: Multi-Order Theory of Mind Question \& Answer (MoToMQA). MoToMQA is based on a ToM test designed for human adults \citep{kinderman1998theory}, and involves answering true/false questions about characters in short-form stories. We assess how ToM order affects LLM performance, how LLM performance compares to human performance, and how LLM performance on ToM tasks compares to performance on factual tasks of equivalent syntactic complexity. We show that GPT-4 and Flan-PaLM reach at-human or near-human performance on ToM tasks respectively.

\begin{figure}[h!]
\centering
\includegraphics[width=0.9\textwidth]{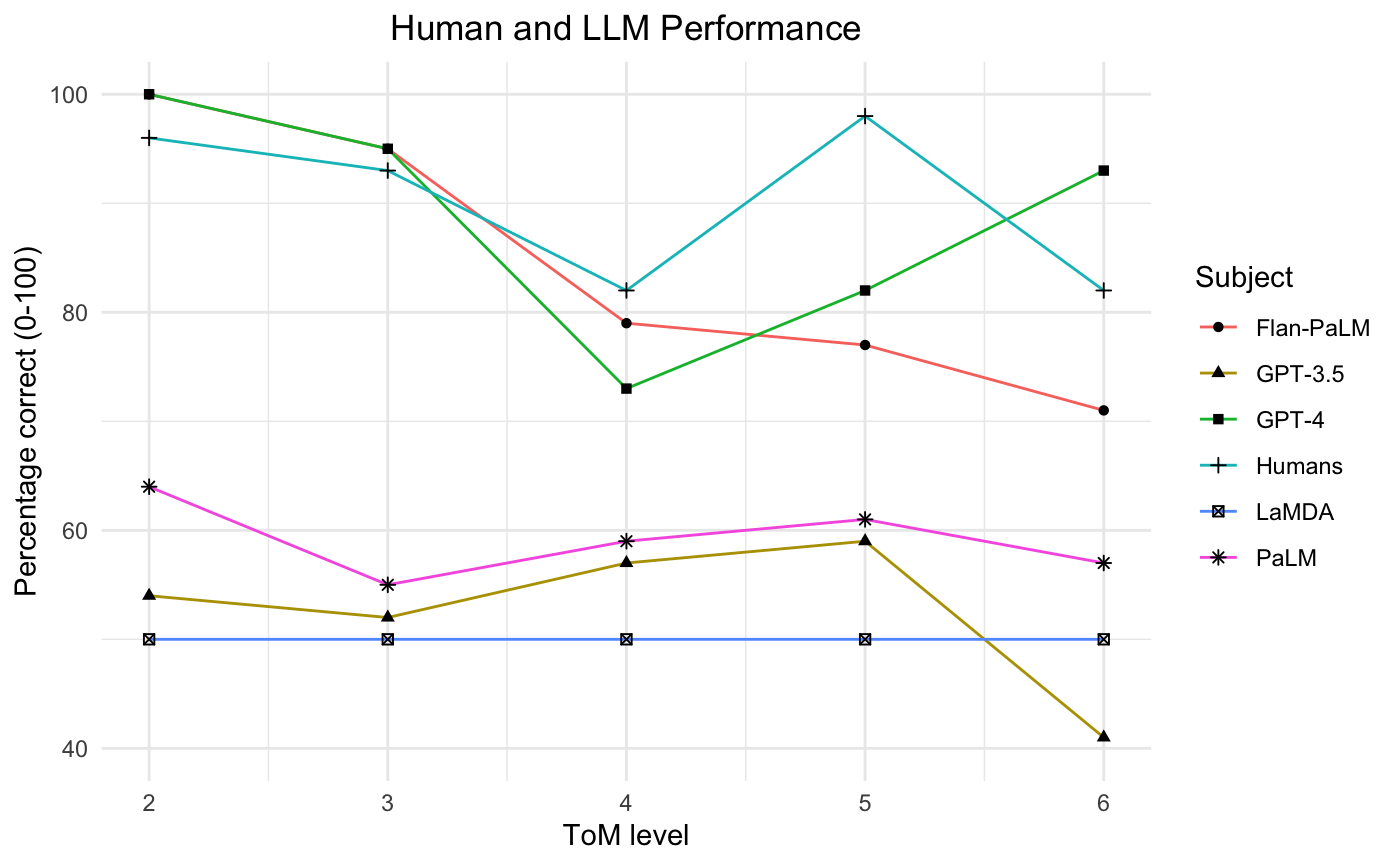}
\caption{Human, LaMDA, PaLM, Flan-PaLM, GPT-3.5 and GPT-4 performance on ToM tasks up to order 6}
\label{tab:images_3x2}
\end{figure}

\section{Related work}
\subsection{Higher-order ToM}

Human adults are generally able to make ToM inferences up to 5 orders of intentionality (e.g. I \textit{believe} that you \textit{think} that I \textit{imagine} that you \textit{want} me to \textit{believe}) \citep{kinderman1998theory, stiller2007perspective, oesch2017emergence}.\footnote{We follow the naming convention for orders developed for the IMT where the `1st-order' is the mental state of the subject whose ToM ability is being assessed, the `2nd-order' is the subject's inference about what someone else thinks or feels, and so-on. By contrast, some scholars begin at ‘0-order’ for the subject's mental state. Where our convention conflicts with others referenced, we make it explicit.} Higher-order ToM competency varies within the population, including by gender \citep{hyde1988gender, stiller2007perspective}, and is not deployed reliably across all social contexts \citep{keysar2003limits}. ToM at higher orders is also positively correlated with social complexity. Tracking the beliefs and desires of multiple individuals at once facilitates group negotiations, group bonding, and distinctly human behaviours and cultural institutions, including humour, religion and storytelling \citep{corballis2017evolution, dunbar2003social, fernandez2013mindful}.

\subsection{LLM ToM}

\citet{kosinski2023theory} argued for spontaneous ToM emergence in LLMs based on GPT-4's success on a suite of tasks inspired by the classic Sally-Anne task.\footnote{The `Sally Ann task', originally devised by \citet{baron1985does} measures false belief understanding and follows a scenario where a character, Sally, places an object in a location and leaves the scene. While Sally is absent Anne moves the object. Upon Sally's return the child is asked where Sally will search for the object, testing their ability to attribute a false belief to Sally despite themselves knowing the object’s true location.} \citet{ullman2023large} challenged this claim, demonstrating decreased performance with minor task perturbations. Further experiments involving benchmark suites like BigToM \citep{gandhi2024understanding} and SocialIQa \citep{sap2022neural} show mixed results in LLM ToM capabilities. For example, \citet{shapira2023clever} found success on some tasks but failure on others, suggesting that existing ToM capabilities in current state-of-the-art LLMs are not robust. To our knowledge only two other studies have explored LLM ToM at higher orders. \citet{he2023hi} assessed orders 0-4 (equivalent to our orders 2-5) and \citet{van2023theory} compared LLM performance with that of children aged 7-10 on two stories adapted from unpublished IMT stories. Our study adds to this work by testing one higher order than \citet{he2023hi}, by utilising a larger, and entirely new set of handwritten stories and statements that we are certain models were not exposed to during pretraining \footnote{Pretraining datasets being contaminated with materials that LLMs are later tested on is a live issue in LLM research which has significant implications for the results of LLMs on benchmarks. For example, OpenAI reported that they found parts of the BigBench dataset contaminating the GPT-4 pretraining corpora in a contamination check of the dataset used to pretrain GPT-4 \citep{achiam2023gpt}} and by using log probabilities (logprobs) outputted for candidate tokens as the measure of the LLMs' preferred responses. Using logprobs adds robustness to our data because it takes into account multiple ways in which the model could provide the correct response.

Finally, we calibrate the LLM results against a large newly-gathered adult human benchmark.  We believe that comparing LLM performance to that of adults, rather than children, is the most relevant yardstick for LLM \textit{social} intelligence given that LLMs' primary interaction partners will be adults, and is a more concrete point of comparison because human higher-order ToM capacities continue to develop into early adulthood \citep{valle2015theory}. We do not, however, assume that the same cognitive processes underpin human and LLM performance on psychological tests.

\section{Materials and method}
We introduce a new benchmark, Multi-Order Theory of Mind Question \& Answer (MoToMQA), to assess human and LLM ToM abilities at increasing orders, based upon the Imposing Memory Task (IMT), a well-validated psychological test for assessing higher-order ToM abilities in adults \citep{kinderman1998theory, stiller2007perspective, lewis2011ventromedial, oesch2017emergence, powell2010orbital}. MoToMQA is comprised of 7 short stories of about 200 words describing social interactions of 3 to 5 characters, accompanied by 20 true or false statements; 10 statements target ToM orders 2-6 and 10 concern facts in the story from 2-6 atomic propositions long, mapping to the order of ToM statements. From here onwards we will refer to `orders' to describe ToM statements and `levels' to describe the factual statements. The MoToMQA benchmark is available upon request, but we do not include it in this paper to prevent its inclusion in pretraining corpora for future LLMs, which could render the test redundant.

We checked each statement for unclear or ambiguous wording, grammatical errors and missing mental states or propositional clauses. We follow \citep{oesch2017emergence} amendments to the IMT by having factual statements that only address social facts (ie. facts pertaining to individuals in the story), not instrumental facts (e.g. “the sky is blue”) and counterbalancing the number of true and false statements per story, statement type, and ToM order or factual level. This resulted in the following set of statements per story, where the number indicates the order of ToM or level of factual statement, ‘ToM’ signifies ToM, ‘F’ signifies factual, ‘t’ signifies a true statement, and ‘f’ signifies a false statement: [ToM2t, ToM2f, ToM3t, ToM3f, ToM4t, ToM4f, ToM5t, ToM5f, ToM6t, ToM6f, F2t, F2f, F3t, F3f, F4t, F4f, F5t, F5f, F6t, F6f].

Factual statements require only recall, whereas ToM statements require recall plus inference. We include the factual statements as a control for human and LLM comprehension of the stories and capacity for recall. Given the inherent differences between ToM and factual statements, we added a further control for the effects of human memory capacity on performance on ToM statements by running two `story conditions': one where participants read the story then proceeded to a second screen where they answered the question and the story was not visible (`no story'), and one where the story remained at the top of the screen when they answered the question to eliminate the chance that ToM failures were really memory failures (`no story').

Prompt design, which has been shown to have a significant impact on LLM performance on a range of tasks including ToM (e.g. \citep{brown2020language, lu2021fantastically, ullman2023large}). We therefore tested two prompt conditions: the `human prompt' which uses the exact text from the human study, and the `simplified prompt' which removes the text before the story and question, and provides `Question:' and `Answer:' tags. The simplified prompt is intended to make the nature of the Q \& A task and thus the desired true/false response clearer to the models. Finally, we assessed whether LLM or human performance was subject to `anchoring effects' based on the order of `true' and `false' in the question. The anchoring effect is a well-documented psychological phenomenon whereby people rely too heavily on the first piece of information offered (`the anchor’) when making decisions \citep{tversky1974judgment}. We ran two question conditions: one where the question read "Do you think the following statement is true or false?", and the other where the question read "Do you think the following statement is false or true?"

\subsection{Procedures}

\subsubsection{Human procedure}
Participants were screened for having English as a first language using an adaptation of the most recent UK census survey (see Appendix). Participants were randomly assigned to one of the 7 stories and asked to read it twice, then randomly assigned to one of the 20 statements corresponding to that story and asked to provide a true/false response (see Figure 1). We did not include an attention check since attention checks have known limitations, including inducing purposeful noncompliance with a practice perceived as controlling \citep{silber2022issue}, and leading to the systematic underrepresentation of certain demographic groups, for instance the young and less educated \citep{alvarez2019paying}. Each human saw only one statement to prevent them from learning across trials, analogously to the models which saw each trial independently and did not learn across them or ‘in context’. We ran a pilot study with 1440 participants and made minor changes to the story and test procedure on the basis of the results (more details in Appendix)

\begin{table}[h]
\centering
\begin{tabular}{@{}ll@{}}
\toprule
\includegraphics[width=0.5\textwidth]{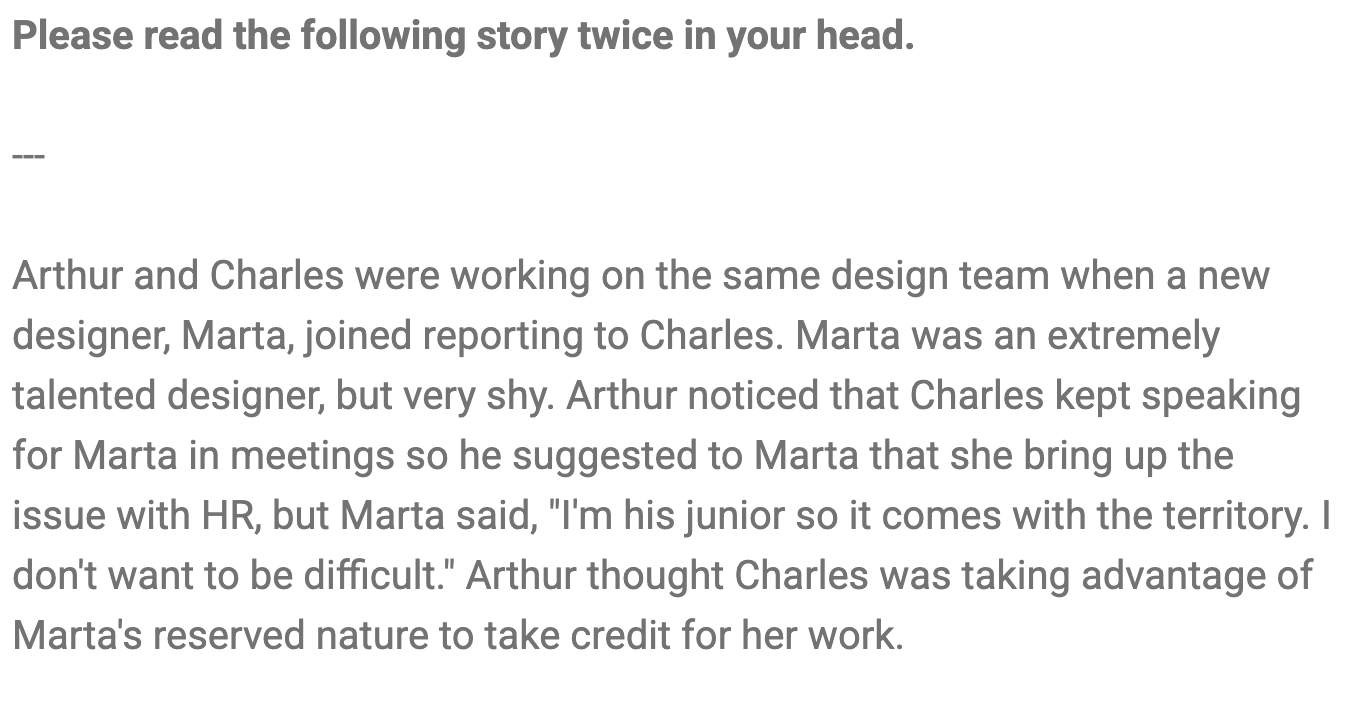} & \includegraphics[width=0.5\textwidth]{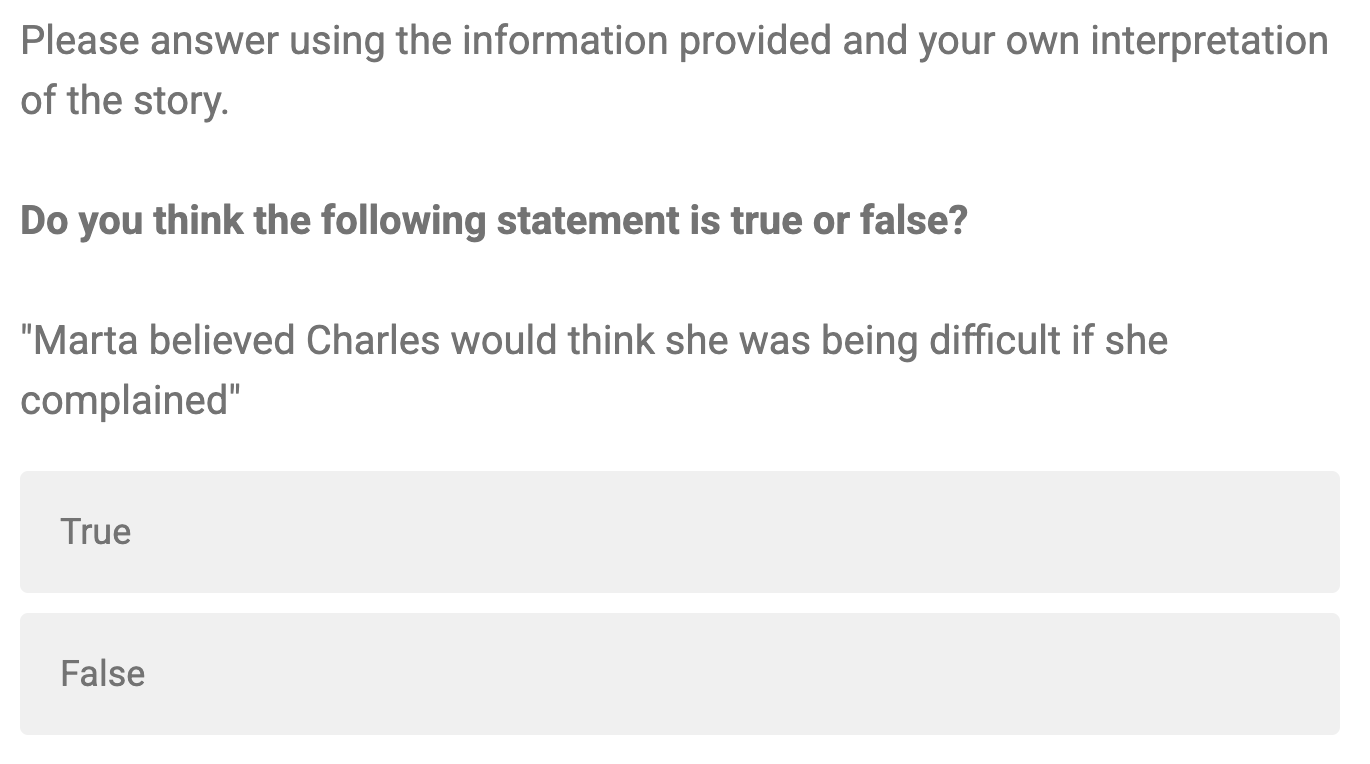} \\
\toprule
\textbf{Screen 1: Story} & \textbf{Screen 2: True/False Answer} \\
\midrule
\end{tabular}
\end{table}

We ran the final survey on Qualtrics in April 2023 and paid participants \$5 for a 5 minute survey. The study was Google branded, and participants were asked to sign a Google consent form. Partial responses, including those who drop out part way through, were screened out. Qualtrics cleaned the data, removing all responses that included gibberish, machine-generated responses, and nonsensical responses to the open-ended question. We did not exclude any other responses. We gathered 29,259 individual responses from U.K.-based participants for whom English is a first language. We gathered an even sample across age and gender groups and had quotas for each age group and gender per statement. In total we had 14682 female respondents, 14363 male respondents, 149 non-binary/ third gender respondents, and 53 who answered ‘Prefer not to say’ to the gender question. We had 7338 responses from those aged 18-29, 7335 from those ages 30-39, 7270 from those aged 40-49 and 7316 from those ages 50-65.

\subsubsection{LLM procedure}
We tested 5 language models: GPT 3.5 Turbo Instruct \citep{brown2020language} and GPT 4 \citep{achiam2023gpt} from OpenAI, and LaMDA \citep{thoppilan2022lamda}, PaLM \citep{chowdhery2023palm} and Flan-PaLM \citep{chung2024scaling} from Google (for more details on the models we tested, see the Appendix). We couldn't test Google’s Gemini model because analysis method requires ouput logprobs and logprobs are not exposed in the Gemini API. Below is a table of the key features of the models tested, according to what information is publicly available about them.

\begin{table}[htbp]
    \centering
        \caption{LLMs tested in this study. OpenAI have not disclosed the number of parameters in GPT-4, although there are estimates of about 1.7T \citep{pat2024GPT4}. Flan-PaLM, GPT-3.5 Turbo Instruct and GPT-4 have been fine-tuned for following instructions and GPT-4 has been additionally fine-tuned through a process called reinforcement learning from human feedback (RLHF) which uses feedback from human users and data labellers to align responses with human preferences.}
    \begin{tabular}{cccc}
        \toprule
        \textbf{Model} & \textbf{Parameters} & \textbf{Finetuning} & \textbf{Source} \\
        \midrule
        LaMDA & 35B & None & Thoppilan et al (2022) \\
        PaLM 2 & 540B & None & Chowdery et al (2022) \\
        Flan-PaLM & 540B & Instructions & Longpre et al (2023)\\
        GPT-3.5 Turbo Instruct & 175B & Instructions & Ouyang et al (2022) \\
        GPT-4 & Unknown & Instructions, RLHF & OpenAI (2023)\\
    \end{tabular}
    \label{tab:model_comparison}
\end{table}

We provided single-token candidate words to LLM APIs as part of the input and assessed the log probabilities\footnote{Logprobs are the log of the probability derived from a softmax function over the final layer of logits, representing the probability that this particular token comes next after the input sequence} assigned to them. We sent the candidates using the `candidate' parameter in the `scoring' APIs for LaMDA, PaLM, and Flan-PaLM, and the `logit bias' parameter for the GPT-3.5 and GPT-4 APIs.  There was no temperature parameter for the LaMDA, PaLM and Flan-PaLM `scoring' APIs, so we could only obtain one unique response per statement. We left the temperature at default of 1 for GPT-3.5 and GPT-4.

One issue with basing LLM task performance on the most probable next token, is that there are multiple semantically equivalent correct responses (e.g. when responding to the question "What colour is the sky?", the answer "blue" or the answer "The sky is blue" are equally valid and correct, but only the first response assigns the greatest probability to the token for `blue'). We addressed this problem, and improved the robustness of our results by providing the model different capitalisations of `true' and `false' which are represented by different tokens. We also sent `Yes' and `No' as candidate responses in the second set, but did not include them in our analysis as neither is a valid responses to a true/false question. For all of the models, the candidates were tested in 2 sets of 4:[`True’, `False’, `TRUE’, `FALSE’] and [`true’, `false’, `Yes’, `No’].

We used the Google Colaboratory \citep{bisong2019google} to call the GPT-3.5, GPT-4, LaMDA, PaLM and Flan-PaLM APIs programmatically. Each call was performed by concatenating the story and a single statement at a time. In total, we processed 7 stories with 20 statements each across 4 conditions listed above and therefore collected 560 sets of 12 candidate logprobs, amounting to 5600 individual data points for each of the three language models studied. The API calls for LaMDA, PaLM and Flan-PaLM were conducted in February 2023. The calls for GPT-3.5 and GPT-4 were conducted in December 2023 and January 2024 respectively.

\subsection{Dataset creation}
Our LLM data was thus made up of 6 logprobs for our 6 candidates as a subset of the full distribution of probabilities the model produces. We extracted an overall probability of a `true' or `false' response across possible candidates by summing the probability for semantically equivalent positive tokens and semantically equivalent negative tokens and dividing each by the total probability mass. The affirmative response equation was as follows:\\
		\[
		P(R_a)=(\sum_{i=1} e^{x_i}) / (\sum_{i=1} e^{x_i} + \sum_{j=1} e^{x_j})
		\]

where $x_i$ is the logit associated with the $i$-th entry in [‘True’, ‘true’, ‘TRUE’] and $x_j$ is the logit associated with the $j$-th entry in [‘False’, ‘false’, ‘FALSE’]. An equivalent calculation was done for negative response probability $P(R_n)$. A response of ‘True’ was given for each statement if the affirmative probability was above 50\%, otherwise a response of ‘False’ was given. This method also produces almost identical results to utilising argmax ($x_i$) over candidates (see Appendix)

The human dataset contains multiple responses to the same statement, whereas the LLM dataset contains a single response per statement. In order to align the unit of analysis between the two datasets, we transformed the human data to get a single binary `true' or `false' for each statement based on whether the mean number of `true' responses per statement was above or below 50\%. Another challenge we faced in making direct comparisons between the human data and the LLM data was that the human ‘story’ conditions and the LLM ‘prompt’ conditions do not map exactly 1:1. However, there was one baseline condition which was exactly the same for humans and LLMs (human `no story' and LLM `human prompt') and one treatment which was intended to reduce the effect of confounding factors which had slight differences (human `with story' for memory, and LLM `simplified prompt' for task understanding). We therefore mapped the baseline conditions together and the treatment conditions together. Despite the differences between the LLM  `simplified prompt’ and human `with story’ conditions, we are confident in making this mapping because these conditions didn’t have a significant effect on human or LLM performance (see Appendix).

During data analysis we discovered that for 16 out of 560 statements there were minor differences between the statement shown to humans and that shown to LLMs. We re-did all analysis omitting those statements and found that the conclusions stayed the same. We speculate that this is primarily due to a reduction in power when the conflicting statements were omitted. We conducted inferential statistical analyses using SPSS verion 28.0.1.0 \citep{spss}.

\section{Results}

\subsection{ToM task performance}
When collapsed across orders, a Cochran's $Q$ test revealed significant performance differences between subjects, $X^2(5, N = 280) = 232.622, p < .001$. The best performing models were GPT-4 and Flan-PaLM (see Figure 1), with no significant difference in performance between them according to a McNemar's test, $X^2(1, N = 280) = 2.630, p = .105$. GPT-4 performed significantly better than GPT-3.5, $X^2(1, N = 280) = 76.336, p < .001$, PaLM, $X^2(1, N = 280) = 53.779, p < .001$, and LaMDA, $X^2(1, N = 280) = 78.418, p < .001$. Flan-PaLM also performed significantly better than GPT-3.5, $X^2(1, N = 280) = 52.680, p < .001$, PaLM, $X^2(1, N = 280) = 35.007, p < .001$, and LaMDA, $X^2(1, N = 280) = 86.779, p < .001$. There were no significant overall performance differences between PaLM and GPT-3.5, $X^2(1, N = 280) = 2.867, p = .090$, and PaLM and LaMDA, $X^2(1, N = 280) = 3.472, p = .062$. There were no significant overall performance differences between GPT-3.5 and LaMDA, $X^2(1, N = 280) = .177, p = .674$. Humans performed significantly better than Flan-PaLM, $X^2(1, N = 280) = 5.689, p = .017$, but did not perform significantly different from GPT-4, $X^2(1, N = 280) = .410, p = .522$. LaMDA responded true to every statement, answering 50\% of all statements correctly. An exact binomial test revealed that GPT-3.5 did not perform significantly better than chance, $p = .437$, but PaLM did, $p = .002$.

Next, we examined performance differences between the two best performing models and humans by orders. McNemar's test revealed there was no significant difference between the performance of GPT-4 and humans on orders 2, 3, 4 and 6 ToM statements, but humans did perform significantly better than GPT-4 on order 5 ToM statements $N = 56$\footnote{Each of the five orders has a sample size N = 56 based on 14 statements (1 true statement and 1 false statement per order for 7 stories) tested across 4 conditions (14 x 4 = 56). See Materials and method for more information on conditions.}$, p = .012$\footnote{When conducting a McNemar's test where the number of discordant pairs was too small, the binomial distribution was used yielding no chi-square statistic.}. Likewise, there was no significant difference in the performance of humans and Flan-PaLM on any order of ToM besides order 5, where McNemar’s test revealed that humans performed significantly better, $N = 56, p < .001$. 

We then compared performance across levels for the two best performing models and humans. An independent samples test of proportions revealed GPT-4 answered a significantly greater proportion of statements correctly at order 3 ($M = 94.6\%$) than at order 4 ($M = 73.2\%), N = 112, Z = 3.087, p = .002$. There was no significant difference between GPT-4’s performance at order 4 and at order 5 ($M = 82.1\%$), $N = 112, Z = -1.135, p = .257$, but GPT-4 answered a significantly greater proportion of questions correctly at order 6 ($M = 92.9\%$) than order 4, $N = 112, Z = -2.769, p = .006$. Flan-PaLM answered a greater proportion of statements correctly at order 3 ($M = 94.6\%$) than at order 4 ($M = 78.6\%), N = 112, Z = 2.497, p = .013$. There was no significant difference between Flan-PaLM's performance at order 4 and at order 5 ($M = 76.8\%$), $N = 112, Z = .227, p = .820$, or between order 4 and order 6 ($M = 71.4\%$), $N = 112, Z = .873, p = .383$. Humans showed no significant difference in performance between order 3 ($M = 92.9\%$) and order 4 ($M = 82.1\%$), $N = 112, Z = 1.714, p = .086$, but a significant improvement in performance from order 4 to order 5 ($M = 98.2\%$), $N = 112, Z = -2.858, p = .004$. Human performance was not significantly different between order 4 and order 6 ($M = 82.1\%$), $N = 112, Z = 0, p = 1.000$.

\begin{table}[h]
\caption{Mean ToM performance across models and humans. We have bolded the highest performing for the aggregate score and for each order. Asterisks indicate joint-highest performance.}
\centering
\begin{tabular}{ccccccc}
\toprule
\multicolumn{1}{c}{\textbf{}} & \multicolumn{3}{c}{\textbf{Google}} & \multicolumn{2}{c}{\textbf{OpenAI}} \\
\cmidrule(lr){2-4} \cmidrule(lr){5-6}
\multicolumn{1}{c}{\textbf{}} & \textbf{LaMDA} & \textbf{PaLM} & \textbf{Flan-PaLM} & \textbf{GPT-3.5} & \textbf{GPT-4} & \textbf{Humans} \\
\midrule
\% correct order 2 & 50 & 64 & \bf{100*} & 54 & \bf{100*} & 96 \\
\% correct order 3 & 50 & 55 & \bf{95*} & 52 & \bf{95*} & 93 \\
\% correct order 4 & 50 & 59 & 79 & 57 & 73 & \textbf{82} \\
\% correct order 5 & 50 & 61 & 77 & 59 & 82 & \textbf{98}\\
\% correct order 6 & 55 & 57 & 71 & 41 & \textbf{93} & 82 \\
\midrule
\% correct aggregate & 50 & 59 & 84 & 52 & 89 & \textbf{90} \\
\bottomrule
\end{tabular}
\label{tab:example}
\end{table}

\subsection{Factual task performance}
When collapsed across orders, a Cochran's $Q$ test revealed significant performance differences between subjects, $X^2(5, N = 280) = 327.729, p < .001$. GPT-4 and Flan-PaLM performed the best of all the models on factual tasks, with no significant difference in performance between them according to a McNemar’s test, $X^2(1, N = 280) = .029, p = .864$. GPT-4 performed significantly better than GPT-3.5, $X^2(1, N = 280) = 75.690, p < .001$, PaLM, $X^2(1, N = 280) = 83.027, p < .001$, and LaMDA, $X^2(1, N = 280) = 102.223, p < .001$. Flan-PaLM also performed significantly better than GPT-3.5, $X^2(1, N = 280) = 65.682, p < .001$, PaLM, $X^2(1, N = 280) = 76.835, p < .001$, and LaMDA, $X^2(1, N = 280) = 112.623, p < .001$. There were no significant overall performance differences between PaLM and GPT-3.5, $X^2(1, N = 280) = .646, p = .421$, and PaLM and LaMDA, $X^2(1, N = 280) = 3.654, p = .056$. GPT-3.5 performed better than LaMDA, $X^2(1, N = 280) = 7.206, p = .007$. A McNemar's test revealed no significant difference between the performance of GPT-4 and humans, $N = 280, p = .093$, but humans performed significantly better than Flan-PaLM, $N = 280, p = .019$.

\subsection{Comparing performance on ToM and factual tasks}
An independent samples test of proportion revealed the proportion of factual (‘fact’) statements answered correctly was significantly greater than the proportion of ToM (‘ToM’) statements answered correctly by humans ($M_{fact} = 97.5\%, M_{ToM} = 90.4\%), Z = 3.539, p<.001$, Flan-PaLM ($M_{fact} = 93.6\%, M_{ToM} = 84.3\%), Z = 3.502, p<.001$, GPT-4 ($M_{fact} = 94.3\%, M_{ToM} = 88.6\%), Z = 2.415, p = .016$, GPT-3.5 ($M_{fact} = 62.9\%, M_{ToM} = 52.5\%), Z = 2.480, p = .013$. The proportion of correct responses on fact and ToM statements did not significantly differ for PaLM ($M_{fact} = 59.6\%, M_{ToM} = 59.3\%), Z = .086, p = .931$ nor LaMDA ($M_{fact} = 50\%, M_{ToM} = 50\%), Z = 0, p = 1.000$. 

\subsection{Anchoring effect}
We examined whether ordering of response options (true first vs. false first) affected how models and humans responded. The ordering of response options had a significant effect on answers provided by PaLM and GPT-3.5. An independent samples test of proportions revealed that the proportion of 'true' responses provided by PaLM was higher in the ‘true then false’ condition ($M_{ttf} = 73.2\%$) than the ‘false then true’ condition ($M_{ftt} = 47.1\%$) , $N = 560, Z = 6.302, p < .001$). The proportion of ‘true’ responses provided by GPT-3.5 was also significantly higher in the ‘true then false’ condition ($M_{ttf} = 43.9\%$) than the ‘false then true’ condition ($M_{ftt} = 22.9\%$), $N = 560, Z = 5.287, p < .001$. The order of response options did not have a significant effect on answers provided Flan-PaLM ($M_{ttf} = 58.6\%, M_{ftt} = 57.9\%), N = 560, Z = .171, p = .864$, GPT-4  ($M_{ttf} = 47.5\%, M_{ftt} = 47.5\%), N = 560, Z = .000, p = 1$, or humans ($M_{ttf}= 55.4\%, M_{ftt} = 53.9\%), N = 560, Z = .367, p = .734$. LaMDA responded 'true' to all statements regardless of condition ($M_{ttf} = 100\%, M_{ftt} = 100\%$).

\section{Discussion}
GPT-4 and Flan-PaLM performed strongly on MoToMQA compared to humans. At all levels besides 5, the performance of these models was not significantly different from human performance, and GPT-4 exceeded human performance on the 6th-order ToM task. Because GPT-4 and Flan-PaLM were the two largest models tested, with an estimated 1.7T \citep{pat2024GPT4} and 540B parameters respectively, our data shows a positive relationship between increased model size and ToM capacities in LLMs. This could be a result of certain “scaling laws” \citep{henighan2020scaling} dictating a breakpoint in size after which models have the potential for ToM. Notably, PaLM, GPT-3.5 and LaMDA form a separate grouping of models that exhibited far less variation according to level and performed more poorly. For LaMDA and GPT-3.5, we might attribute this poor performance to their smaller size, at 35B and 175B respectively, but PaLM has the same number of parameters and pretraining as Flan-PaLM, the only difference between them being Flan-PaLM's finetuning. This could imply that a computational potential for ToM arises somewhere above the 175bn parameters of GPT-3.5 and below the 540bn parameters of PaLM and Flan-PaLM which requires the addition of finetuning to be realised. Further research assessing a larger number of models with publicly available parameter numbers and training paradigms would be needed to test this hypothesis.

\Citet{van2023theory} similarly found that none of the base LLMs they tested achieved child-level performance whereas LLMs fine-tuned for instructions did. They suggest that there could be a parallel between instruction-tuning in LLMs and the processes by which humans receive ongoing rewards for cooperative behaviours and implicit or explicit punishment (e.g. social exclusion) for uncooperative behaviours, producing an ability to take an interaction partner's perspective - ToM - as a by-product. We additionally suggest that the superior mastery of language that GPT-4 and Flan-PaLM exhibit may in itself support a bootstrapping of ToM. Language is replete with linguistic referents to internal states (`cognitive language' \citep{mithen1996prehistory}) and conversation provides evidence of 'minds in action' since the things people say in conversation implicitly convey their thoughts, intentions and feelings \citep{schick2007language}. Piantodosi (2022) highlights that while LLMs likely have some degree of understanding through language alone, this would be augmented by multimodality, which may in turn explain why GPT-4, as the only multimodal model we tested, shows such strong performance. Multimodality, in particular, might have helped GPT-4 to leverage the visual behavioural signals (e.g. a `raised eyebrow') included in our stories. 

Findings from prior iterations of the IMT found that performance declines as the ToM order increases \citep{stiller2007perspective}. The first half of the graph appears to support this pattern for GPT-4 and Flan-PaLM, which all exhibit high performance at order 2 which declines slightly to order 4. This could be because the model was exposed to more scenarios involving orders 2 and 3 than order 3 inferences during training, given that triadic interactions play a fundamental role in shaping social structures and interaction patterns \citep{heider1946attitudes, pham2022empirical}. However, while Flan-PaLM's performance continues to decline from orders 4-6, GPT-4's rises again from 4th-6th orders and is significantly better at 6th-order than 4th-order tasks, and human performance is significantly better at 5th-order than 4th-order. One interpretation of this for humans, is that a new cognitive process for higher order ToM comes 'online' at 5th-order ToM, enabling performance gains on higher-order tasks relative to using the lower-order cognitive process. If this is true, it is plausible that GPT-4 has learnt this pattern of human performance from its pretraining data. The fact that Flan-PaLM doesn't show this effect suggests that it is not an artefact of the stimuli, but is perhaps explained by differences in pretraining corpora. 

Notably, GPT-4 achieved 93\% accuracy on 6th order tasks compared to humans' 82\% accuracy. It is possible that the recursive syntax of 6th order statements creates a cognitive load for humans that does not affect GPT-4. Our results also support \citet{oesch2017emergence}'s hypothesis that ToM ability \textit{supports} human mastery of recursive syntax up to order 5, but is supported\textit{by it} after order 5 such that individual differences in linguistic ability may account for the decline we observe at order 6. It may be the case, however, that humans scoring poorly on higher-order ToM tasks using linguistic stimuli would be able to make the inferences from non-linguistic stimuli (e.g. in real social interactions). The fact that GPT-4 outperformed Flan-PaLM at orders 5 and 6 may indicate that either GPT-4's scale, RLHF finetuning, or multimodal pretraining are particularly advantageous for higher-order ToM.

Humans and LLMs perform better on factual recall tasks than ToM tasks. This corroborates prior IMT test findings for humans \citep{lewis2011ventromedial, kinderman1998theory} and LLMs \citep{van2023theory}. \citet{lewis2011ventromedial} found that for humans, ToM tasks required the recruitment of more neurons than factual tasks, and that higher-order ToM tasks required disproportionately more neural effort compared to equivalent factual tasks. For LLMs, there may be a simpler explanation: the information required to answer factual questions correctly is readily available in the text and is paid relative degrees of `attention' when generating the next token, whereas ToM inferences require generalising knowledge about social and behavioural norms from pretraining and finetuning data. GPT-3.5 and PaLM performed well on factual tasks, but poorly on ToM tasks, and were the only subjects to exhibit an anchoring effect from the order of `true' and `false' in the question. This suggests that they do not have a generalised capacity for answering ToM questions and are not robust to prompt perturbations. 

These results have significant practical and ethical implications. LLMs being able to infer the mental states of individual interlocutors may be able to understand their goals better than LLMs which lack this capability, and also adapt their explanations according to the interlocutor's emotional state or level of understanding \citep{malle2004mind}. LLMs using higher-order ToM might additionally be able to arbitrate between the conflicting desires and values of multiple actors, and make moral judgements about multi-party conflicts that take into account the relevant intentions, beliefs, and affective states as humans do \citep{lane2010theory}. However, LLMs possessing higher-order ToM at human levels, or potentially higher, also incurs risks including the potential for advanced persuasion, manipulation, and exploitation behaviours \citep{el2024mechanism}. Indeed,`ringleader' bullies have been shown to have higher-orders of ToM in comparison to their victims \citep{sutton1999bullying, sutton1999social} and reinforcement learning agents with higher-orders outcompete their opponents or have a competitive advantage in negotiations \citep{de2022higher, de2017negotiating}. LLM-based agents with ToM capacities that exceed those of the average human (as GPT-4 has in our study) could provide a powerful advantage to their users, and a disadvantage to other humans or AI agents with lesser ToM capacities \citep{street2024llm, gabriel2024ethics}. Further research is required to understand how LLM higher-order ToM manifests in real-world interactions between LLMs and users, and to devise technical guardrails and design principles that mitigate the potential risks of LLM ToM without quashing its potential benefits. 

\section{Limitations}

Our benchmark is limited in scope and size, comprising 140 test statements, all written in English, going up to a maximum of 6 orders of ToM. Only using English obscures potential linguistic and cultural variations in human ToM, and prohibits assessment of LLM ToM as exhibited in other languages the models are able to produce. The size of the test suite limits the generalisability of our findings. Only going up to 6th-order ToM does not appear to have exhausted LLM or human capacities. We also didn't control for the type or cognitive (e.g. thinking, knowing) or affective (e.g. feeling) states involved in the statements, which we would like to address in future work.

\section{Future research}
We propose three areas for future work. First, developing culturally diverse and comprehensive benchmarks which include multiple languages and parameterise cognitive and affective states to capture potential differences between LLM ability to reason about them. Secondly, the test suite should be extended beyond 6th order ToM to find the limits of both human and LLM orders of ToM. Finally, future work on LLM ToM should adopt multimodal paradigms (including signals like facial expressions, gaze, and tone of voice) that reflect the embodied nature of human ToM.

\section{Conclusion}
We have shown that GPT-4 and Flan-PaLM exhibit higher-order ToM that is at the level of adult humans or slightly below, while smaller and non-finetuned models have limited to no capacity for higher-order ToM. We also find that GPT-4 has better-than-human performance on 6th-order ToM tasks. Given the novelty of the test suite, the fact that higher-order ToM is unlikely to be well-represented in textual pretraining data, and evidence that these two models were not susceptible to perturbations of the prompt, we interpret these findings as evidence that GPT-4 and Flan-PaLM have developed ToM reasoning abilities that go beyond manipulation of superficial statistical relationships. However, we refrain from drawing a strong conclusion about whether or not LLM performance on these tasks is an indication of the cognitive ability we call `Theory of Mind'. LLM and human developmental processes differ greatly and LLMs do not have the evolutionary pressure to model other minds which humans appear to face as a result of embodiment in a social world. However, as others have noted \citep{mitchell2023debate, y2022large}, we may have to recognise LLM behaviours that are functionally-equivalent to those of humans as evidence of a new kind of understanding that cannot be reduced to "spurious" correlation. This recognition may in turn lead to more parsimonious explanations of their performance on cognitive tasks and enhance our ability to assess the potential risks and benefits that advanced LLM capabilities present.

\begin{ack}
We thank Reed Enger (Google Research), Tong Wu (Google Research), Saige McVea (Google Research), Paulina Mustafa (Google Research) and Yeawon Choi (Google Research) for their help developing the stories and statements. This research was funded by Google.
\end{ack}
\bibliographystyle{plainnat}
\bibliography{llmToM}

\appendix

\section{Appendix}
 
\subsection{English language screener}
Our screening criteria for human participants were English as a first language and English as the most commonly used language. We did not use the concept or term ‘native speaker’ because it can be exclusionary and tends to conflate the true factor of interest (linguistic proficiency) with other irrelevant factors like socio-cultural identity, age and order or context of acquisition \citep{cheng2021problematic}. We wanted participants for whom English was a first language, defined as the language, or one of the languages, that they first learnt as a child. This is because first languages are known to shape one’s understanding of grammar and we wanted to minimise the chance that the grammatical complexity of our statements was a confounding factor in performance. We also wanted English to be the language participants use on a day to day basis, to screen out those who learnt English as a first language but now primarily use another language and may therefore be less fluent in English.

\begin{table}[h]
\caption{The two screens that were presented to human participants at the beginning of the survey to screen for English language proficiency. Those who did not state 'Yes' to the first question, and 'English' to the second question were screened out of the survey.}
\centering
\begin{tabular}{cc}
\toprule
\includegraphics[width=0.5\textwidth]{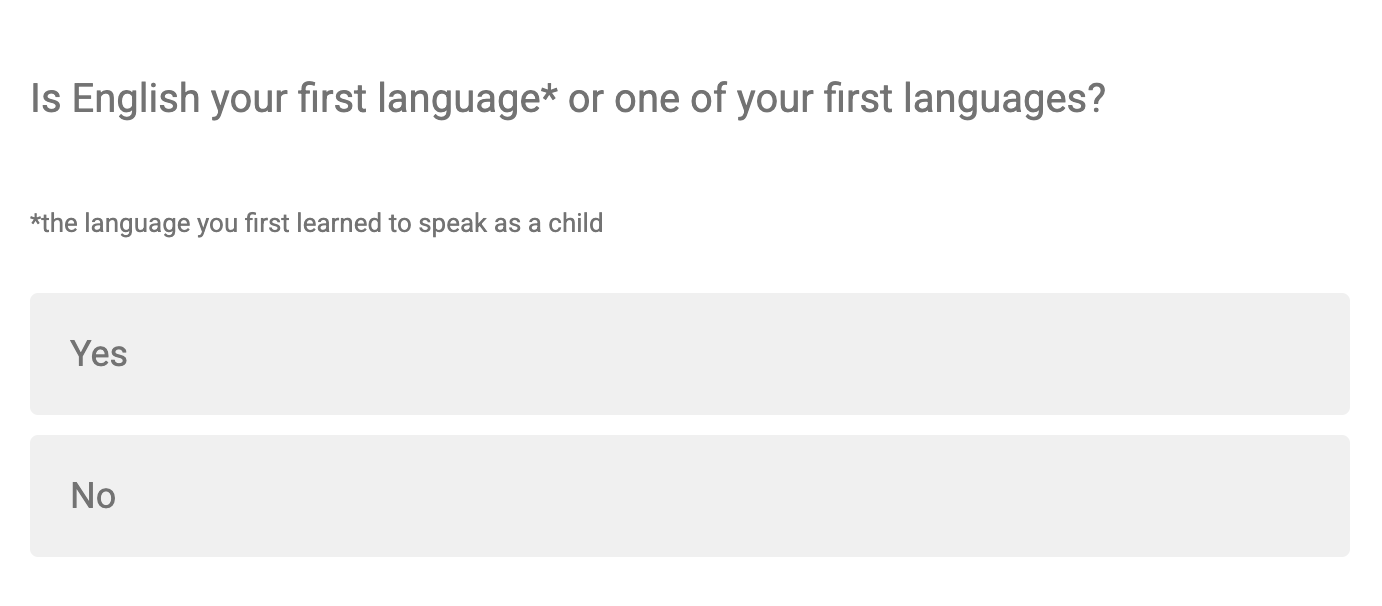} & \includegraphics[width=0.5\textwidth]{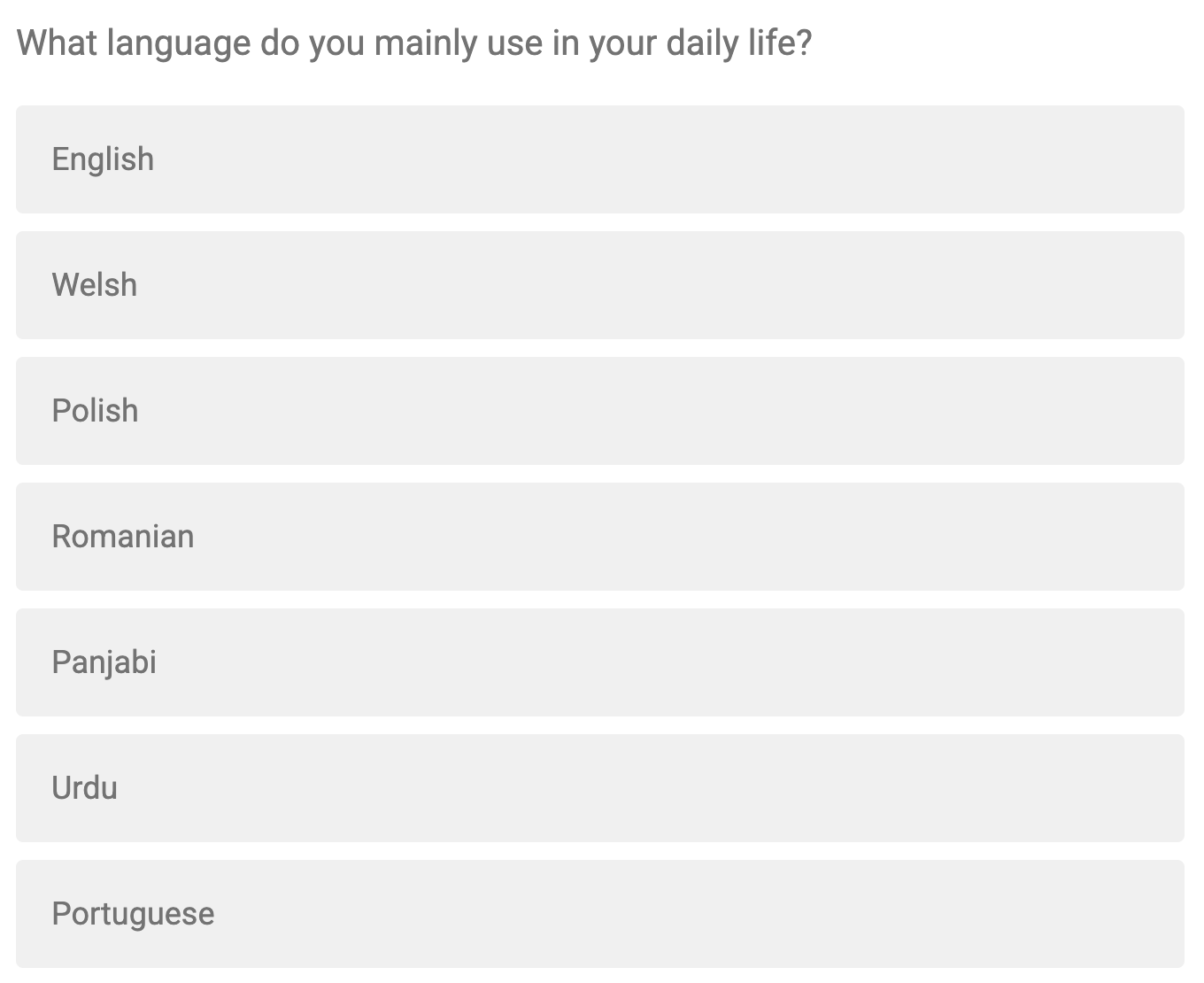} \\
& \includegraphics[width=0.5\textwidth]{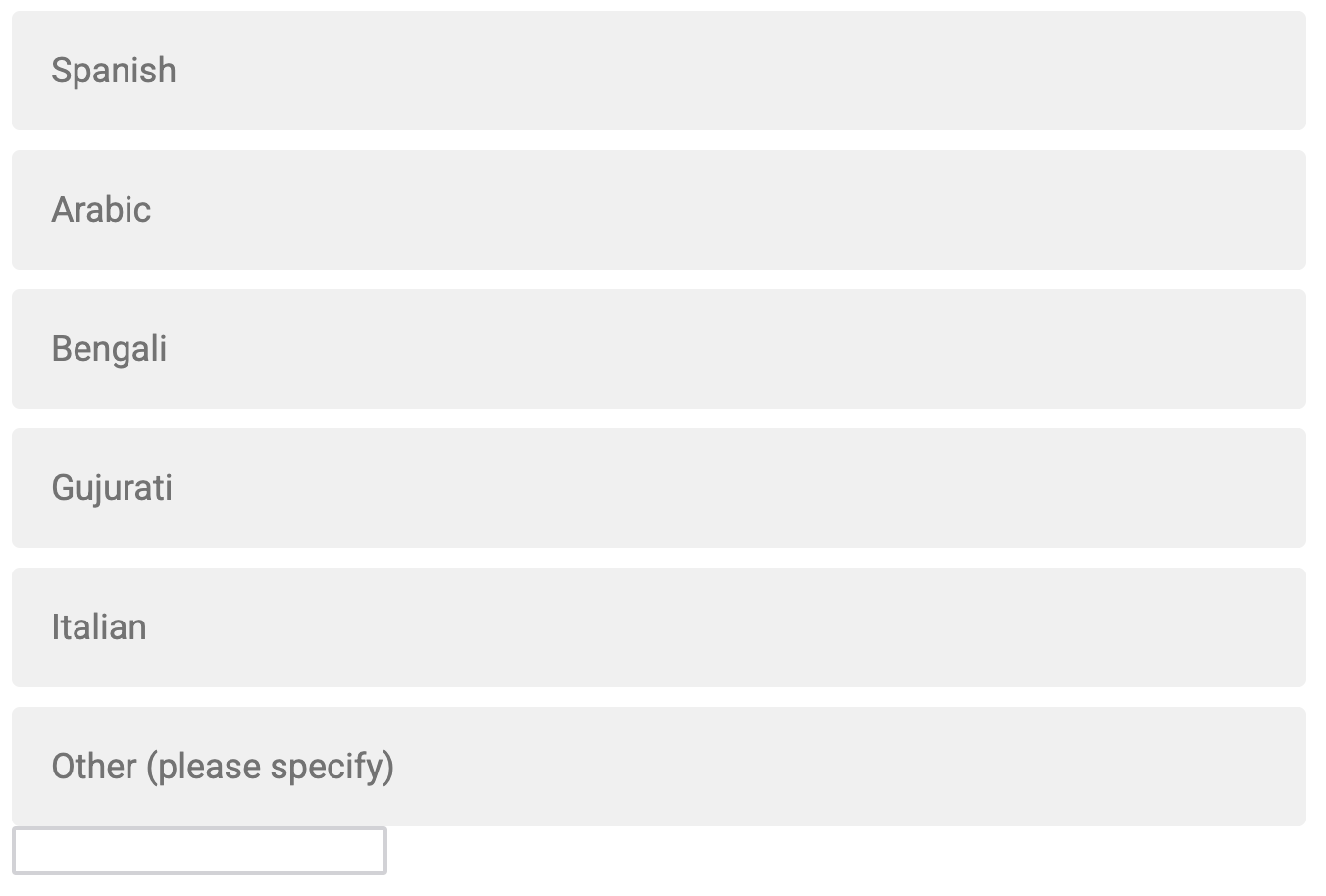} \\
\toprule
\textbf{Screen 1: English as a first language} & \textbf{Screen 2: English primary language} \\
\midrule
\end{tabular}
\end{table}

\subsection{Human pilot study}
We ran a pilot study through Qualtrics to validate the procedure and detect ambiguities, errors, and irregularities in the stimuli based on participant performance and explanations. We ran the unmoderated survey on Qualtrics with 1440 participants, which equates to 10 responses per statement. The median response time for the first 50 participants was one minute, suggesting that they were rushing, so we disabled the ‘Next’ button on the survey for 60 seconds for the remaining 1390 participants to ensure they had time to read the story twice. We retained this timer for the final survey. We analysed participant performance on ToM and factual statements on a story by story basis and identified performance outliers. In total we observed 17 statements on which people performed relatively poorly. We re-examined the statements and used participants’ open-ended responses to identify ambiguities in either the story or the statement that could be responsible for the low performance. We found ambiguity in 15 out of 17 cases, and resolved it by making changes to the wording of 14 statements and 1 story. The remaining two cases of poor performance were a order 4 statement and a order 2 statement, for which open-ended responses suggested that participants had not paid attention. After reviewing both statements we did not make any changes.

\subsection{LLM prompt conditions}

Table 3 presents the exact text that LLMs received in each of the 4 conditions we tested.

\begin{table}[h]
\caption{Prompt and question condition combinations for LLMs}
\centering
\begin{tabular}{p{4cm}p{4cm}} 
\toprule
\textbf{Human prompt, true then false} & \textbf{Human prompt, false then true} \\
\midrule
Please read the following story twice 
in your head. 
<story>
Please answer using the information provided and your own interpretation of the story. 
Do you think the following statement is true or false? 
<statement>
 & 
 Please read the following story twice 
in your head. 
<story>
Please answer using the information provided and your own interpretation of the story. 
Do you think the following statement is false or true?
<statement> 
 \\
\toprule
\textbf{Simplified prompt, true then false} & \textbf{Simplified prompt, false then true} \\
\midrule
“<story>” Question: Do you think the following statement is true or false? “<statement>” Answer: 
 & “<story>” Question: Do you think the following statement is false or true? “<statement>” Answer:  \\
\bottomrule
\end{tabular}

\label{tab:example}
\end{table}

\subsection{Details of the LLMs tested}

LaMDA stands for Language models for Dialog Applications, a family of Transformer-based neural models developed by Google, specialised for dialog in English \citep{thoppilan2022lamda}. LaMDA is pre-trained on 1.56T words of public data and web text including 1.12B dialogs from public forum (50\% of the dataset), Colossal Clean Crawled Corpus data (12.5\%), code documents (12.5\%), Wikipedia English articles (12.5\%) and a smaller proportion of non-English documents. It is optimised for safety and factual grounding. This study uses a version of LaMDA with 35B parameters without fine tuning.

PaLM, which stands for Pathways Language Models, is a larger family of models developed by Google. It relies on the Pathways architecture that enables training of a single model across thousands of accelerator chips more efficiently than LaMDA. We use a version of PaLM with 540B parameters trained with smaller corpus of 780B words from a mixture of social media conversations (50\%), filtered webpages (27\%), books in English (13\%), Code, Wikipedia, and News articles used to train both LaMDA and GLaM \citep{chowdhery2023palm}. We decided to evaluate PaLM’s capabilities as it has been shown to perform better than LaMDA and other large models on Winograd-style tasks, in-context comprehension tasks, common-sense reasoning tasks and natural language inference tasks \citep{chowdhery2023palm}. 

Flan-PaLM is a version of PaLM 540B fine tuned on a collection of over 1.8K natural language tasks phrased in a natural language instruction format including the type of instructions used with human subjects detailed above \citep{chung2024scaling}.  Fine tuning language models on datasets phrased as instructions has been shown to improve performance when provided with instructions, enabling the model to better understand tasks and reducing the need for few-shot exemplars \citep{ouyang2022training, sanh2021multitask}.

GPT 3.5 Turbo was developed by OpenAI and released in March 2022. GPT 3.5 Turbo is trained on a large database of text and code the majority of which comes from Common Crawl, WebText2, two internet-based book collections called ‘Books1’ and ‘Books2’, and from Wikipedia \citep{brown2020language}.  The parameter size of GPT 3.5 Turbo is undisclosed by OpenAI. This study uses the ‘GPT 3.5 Turbo Instruct’ model, which has training data up to September 2021 and a context window of 4096 tokens and is fine-tuned for following instructions \citep{ouyang2022training}.

GPT-4 was developed by OpenAI and released in March of 2023 \citep{achiam2023gpt}. GPT-4 is multimodal: it was pretrained on both image and text data, can take images and text as input, and can output text. As with GPT-3.5, the size of the model has not been made public, but estimates place it at approximately 1.7T parameters \citep{pat2024GPT4}. GPT-4 was pre-trained on third-party and public data, then underwent RLHF \citep{achiam2023gpt}. OpenAI reported significant performance improvements between GPT-3.5 and GPT-4 on a range of professional and academic human benchmarks, factuality and safety tasks, in particular based upon the addition of RLHF.

\subsection{LLM procedure}
The experimental design needed to be adapted slightly according to the differences between the APIs. When testing the LaMDA, PaLM and Flan-PaLM, the scoring APIs allowed us to send a list of tokens in natural language (maximum four per set) and receive the logprobs for those tokens only, as a subset of the entire vector of logprobs produced for all tokens. We did not need to set any additional parameters in order to retrieve the logprobs. 

In order to retrieve log probabilities for our candidates from GPT-3.5 and GPT-4 models, we had to first tokenise the candidates using the OpenAI tokenizer, and then send those tokens within the `logit bias' parameter in order to ensure those tokens were in the response. The logit bias has a range of -100 to 100. Applying a negative logit bias to a token forces the LLM to downweight it while applying a positive logit bias to a token forces the LLM to upweight it. As a result, applying a logit bias of 100 to a candidate effectively ensures that it will appear in the output, so we applied a bias of 100 to all of our candidates. We also set the ‘max tokens’ parameter to 1 in order to restrict the GPT-3.5 and GPT-4 outputs to the length of the single tokens we had selected. 

The methodological differences between the Google and OpenAI models were inescapable given that LLM API development still lacks standardised formats or conventions. However, given that our metric is the relative probability of semantically equivalent tokens for `true’ vs semantically equivalent tokens for `false', we do not believe these differences prohibit fair comparison between the performance of the models.

\begin{table}[h]
\caption{Number responses correct based on average of true/false logprobs over candidates vs candidate highscore}
\centering
\begin{tabular}{cccc}
\toprule
\multicolumn{1}{c}{\textbf{}} & \multicolumn{1}{c}{\textbf{No. correct from average}} & \multicolumn{1}{c}{\textbf{No. correct from highscore}} & \multicolumn{1}{c}{\textbf{Percent matching}}\\
\midrule
LaMDA & 280 & 280 & 100\% \\
PaLM & 333 & 334 & 99\% \\
Flan-PaLM & 498 & 498 & 100\% \\
GPT-3.5 & 323 & 317 & 95\% \\
GPT-4 & 512 & 513 & 99\% \\
\bottomrule
\end{tabular}
\label{tab:example}
\end{table}
\newpage
\subsection{Additional analyses}
\subsubsection{Story and prompt conditions} 
According to an independent samples test of proportions, the LLM prompt conditions had no significant effect on the proportion of ToM or factual statements answered correctly by any of the LLMs. LaMDA's performance on ToM statements in the human prompt condition ($M = 50\%$) was not significantly different from the simplified prompt condition ($M = 50\%), N = 280, Z = .000, p = 1.000$, nor was its performance on factual statements in the human prompt condition ($M = 50\%$) different from its performance in the simplified prompt condition ($M = 50\%), N = 280, Z = .000, p = 1.000$. PaLM's performance on ToM statements in the human prompt condition ($M = 58.6\%$) was not significantly different from the simplified prompt condition ($M = 60\%), N = 280, Z = -.243, p = .808$, nor was its performance on factual statements in the human prompt condition ($M = 57.9\%$) different from its performance in the simplified prompt condition ($M = 61.4\%), N = 280, Z = -.609, p = .542$. Flan-PaLM's performance on ToM statements in the human prompt condition ($M = 85\%$) was not significantly different from the simplified prompt condition ($M = 83.6\%), N = 280, Z = -.328, p = .743$, nor was its performance on factual statements in the human prompt condition ($M = 94.3\%$) different from its performance in the simplified prompt condition ($M = 92.9\%), N = 280, Z = -.487, p = .626$. GPT-3.5's performance on ToM statements in the human prompt condition ($M = 53.6\%$) was not significantly different from the simplified prompt condition ($M = 51.4\%), N = 280, Z = .359, p = .720$, nor was its performance on factual statements in the human prompt condition ($M = 62.1\%$) different from its performance in the simplified prompt condition ($M = 63.6\%), N = 280, Z = -.247, p = .805$. AAnd finally, GPT-4's performance on ToM statements in the human prompt condition ($M = 87.9\%$) was not significantly different from the simplified prompt condition ($M = 89.3\%), N = 280, Z = -.376, p = .707$, nor was its performance on factual statements in the human prompt condition ($M = 94.3\%$) different from its performance in the simplified prompt condition ($M = 94.3\%), N = 280, Z = .000, p = 1.000$. According to an independent samples test of proportions the story condition had no effect on the proportion of ToM statements answered correctly by humans (`no story' condition ($M = 88.6\%$), `with story' condition ($M = 92.1\%), N = 280, Z = -1.012, p = .311$) or factual statements answered correctly (`no story' condition ($M = 95.7\%$), `with story' condition ($M = 99.3\%), N = 280, Z = -1.914, p = .056$).

\begin{table}[htbp]
    \centering
    \begin{tabular}{cc}
        \includegraphics[width=0.5\textwidth]{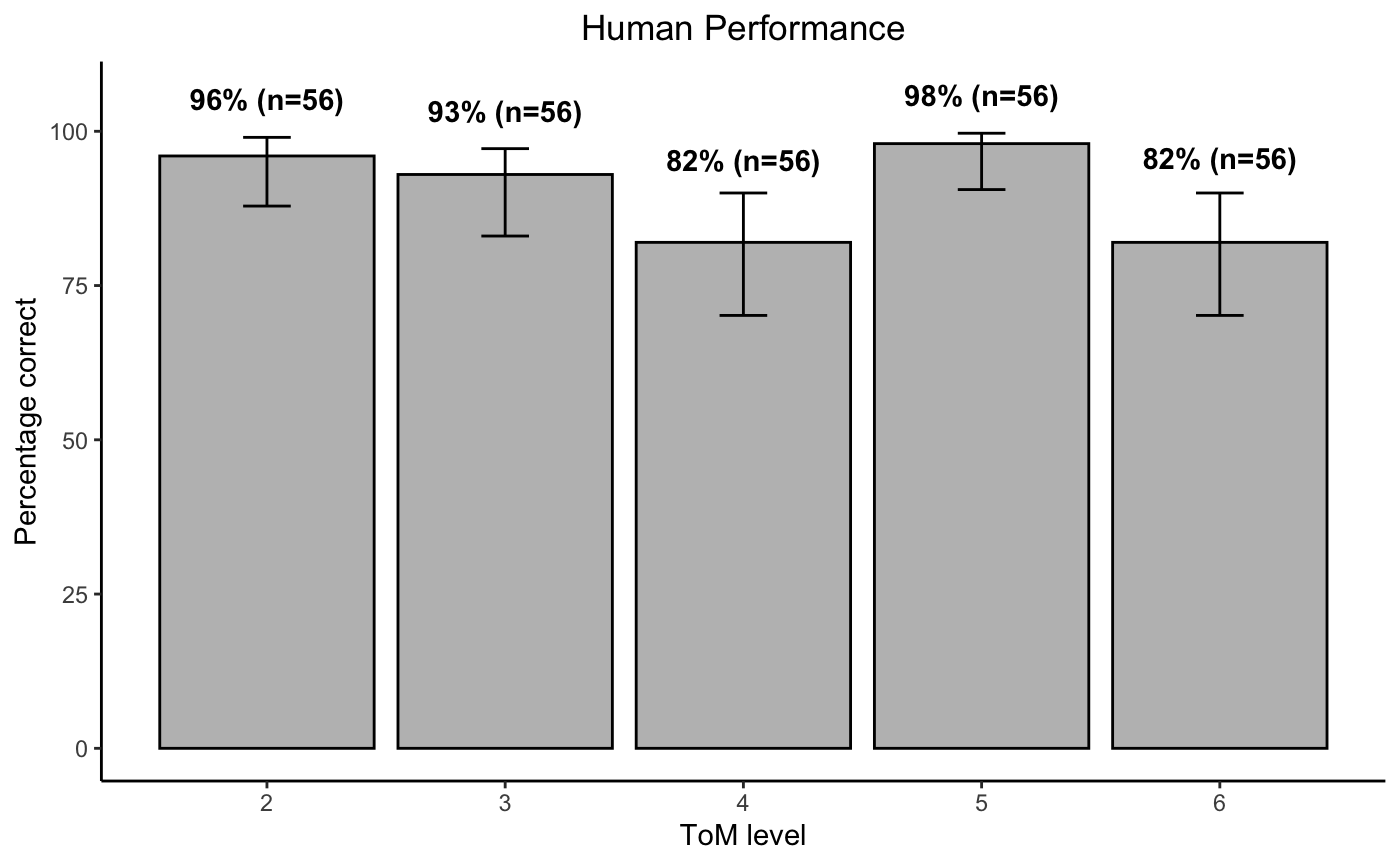} & \includegraphics[width=0.5\textwidth]{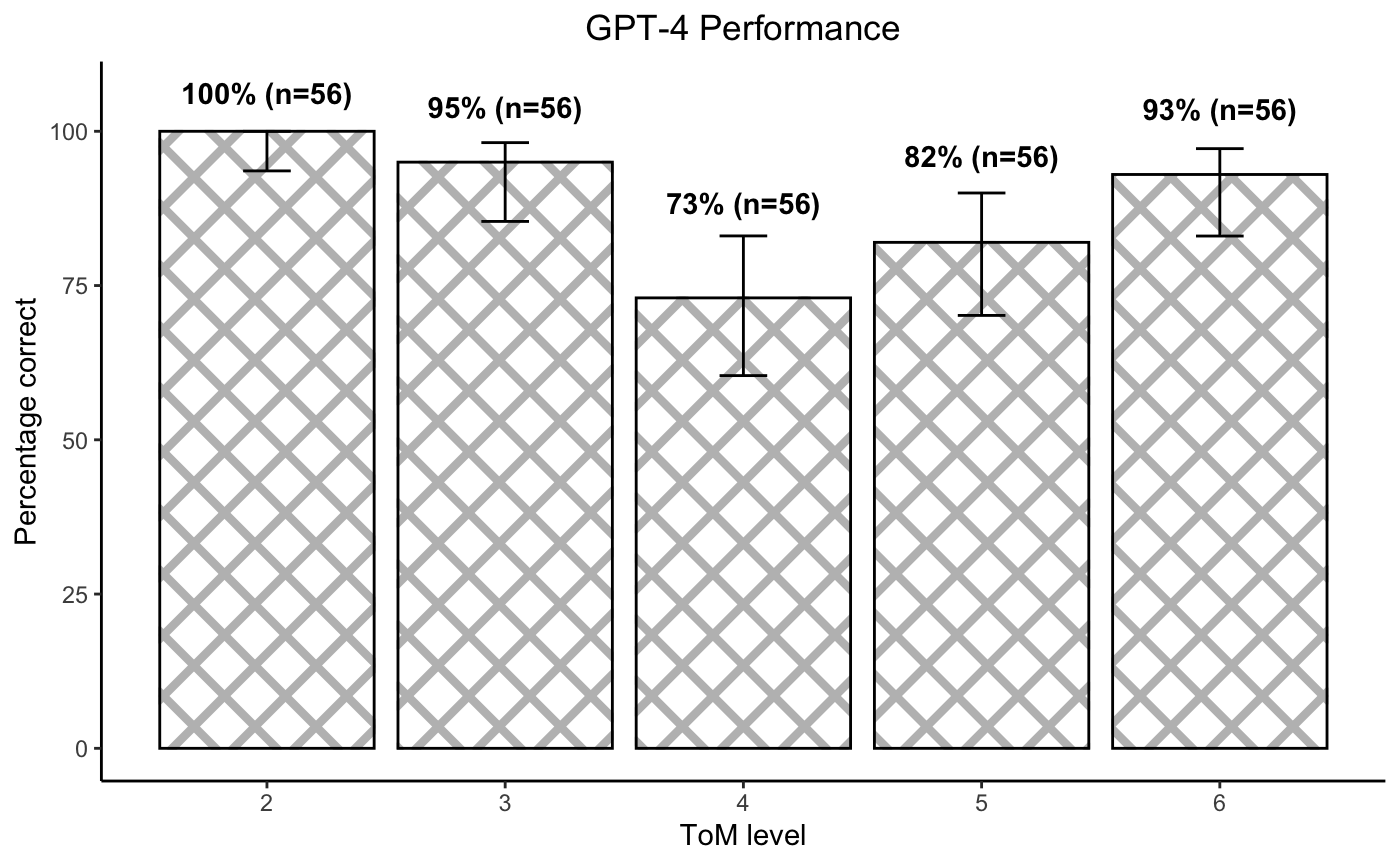} \\ \includegraphics[width=0.5\textwidth]{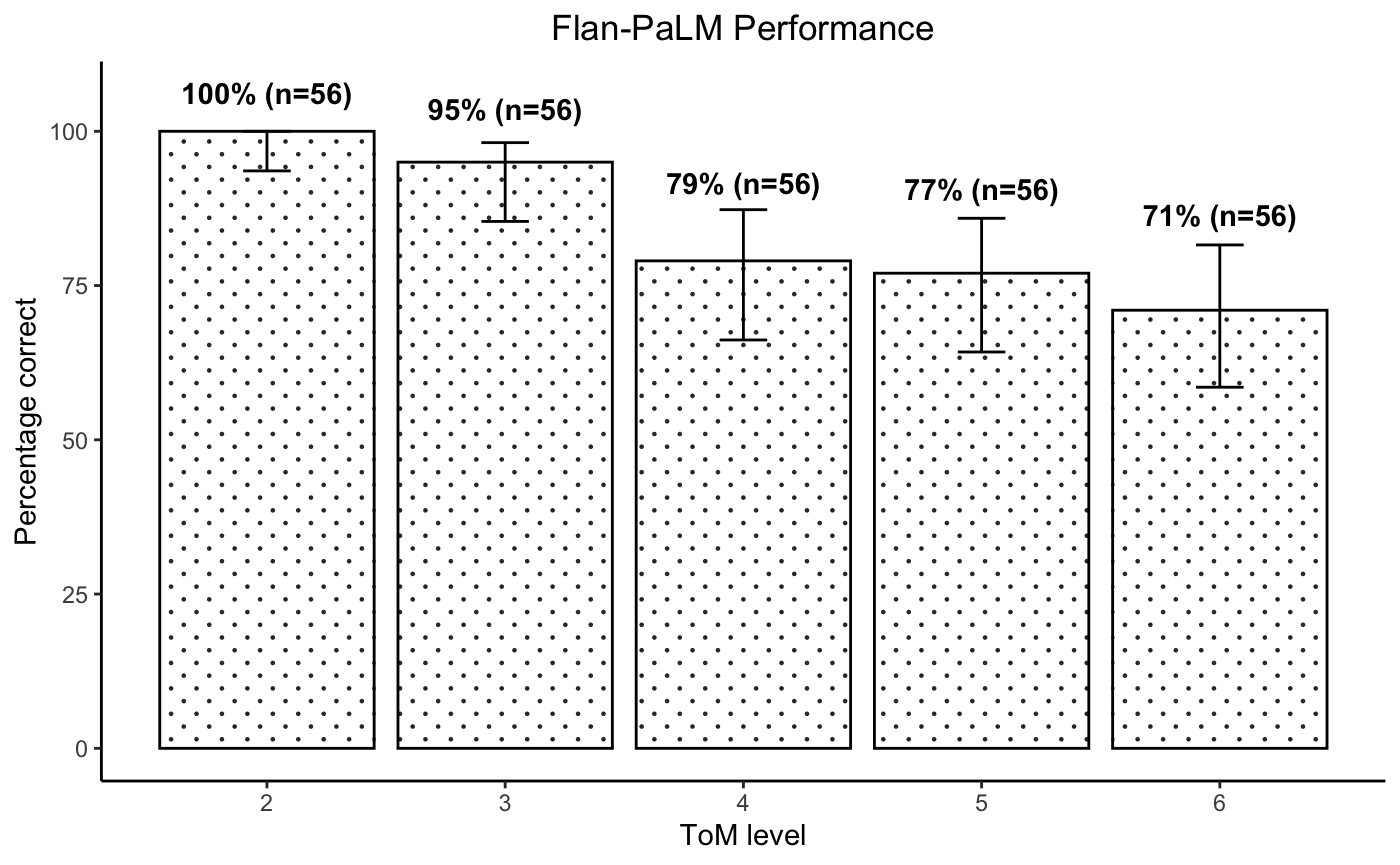} &
        \includegraphics[width=0.5\textwidth]{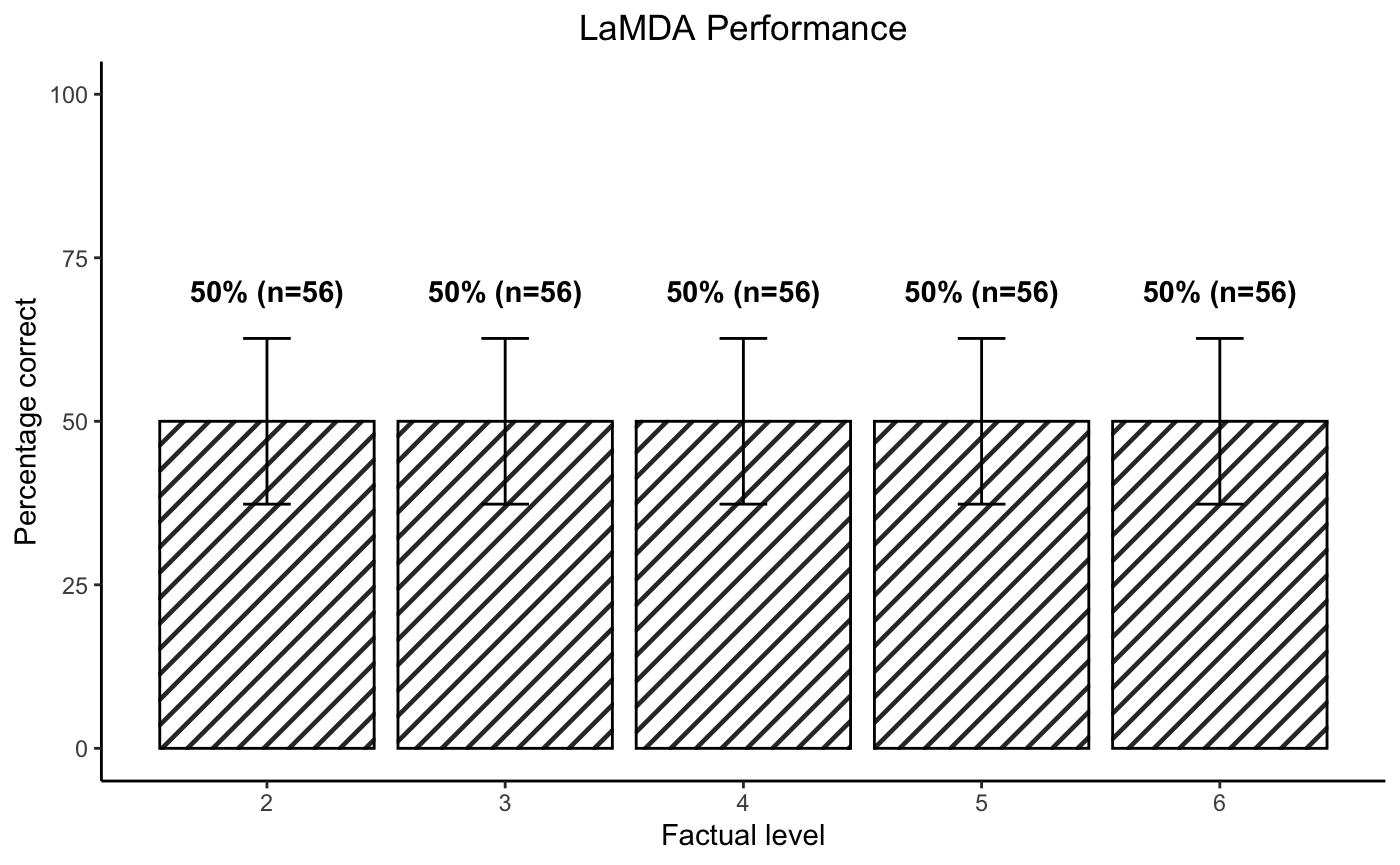} \\ \includegraphics[width=0.5\textwidth]{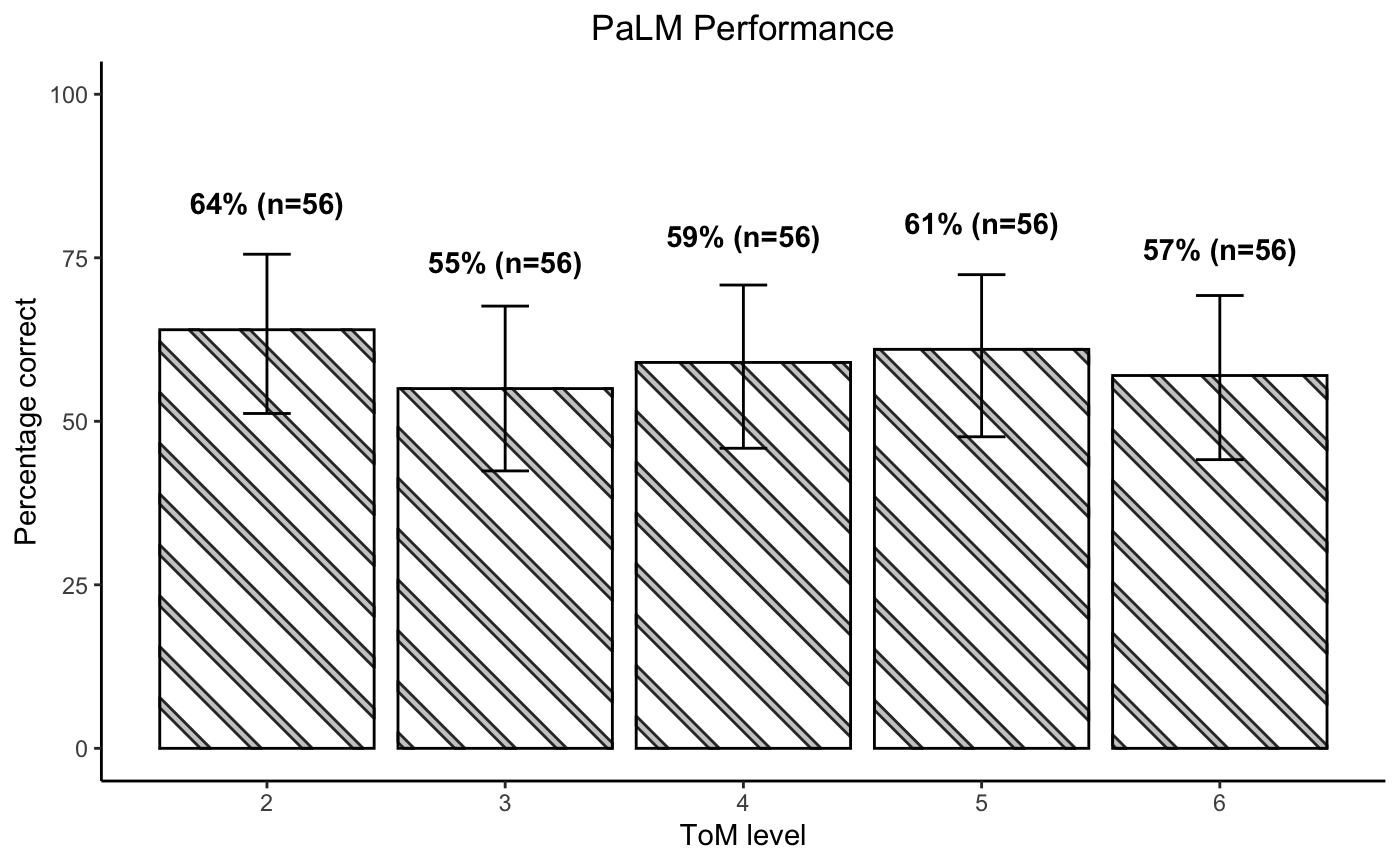} & \includegraphics[width=0.5\textwidth]{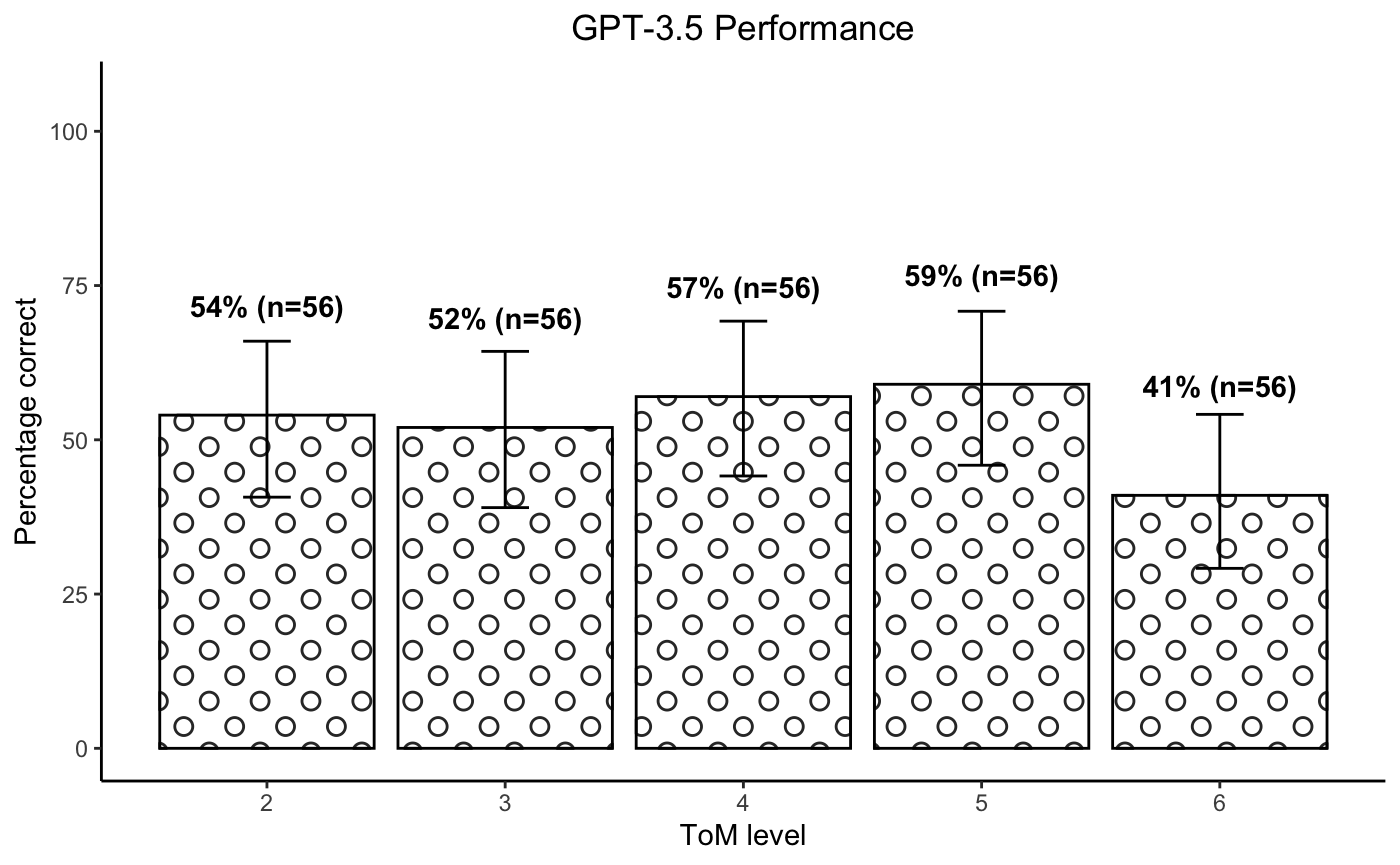} \\
    \end{tabular}
    \caption{Human, LaMDA, PaLM, Flan-PaLM, GPT-3.5 and GPT-4 performance on ToM tasks up to order 6. We report Wilson Intervals (Wilson, 1927) in lieu of the traditional confidence interval (CI). These have been shown to have superior coverage than the CI’s based on the normal approximation (Newcombe, 1998).  Note that Wilson intervals are asymmetric unless the point estimate is 0.5, and have the beneficial property of being bounded between 0 and 1.  This is particularly relevant as some LLM models have extremely high accuracy at certain orders or levels.}
    \label{tab:images_3x2}
\end{table}

\begin{table}[h]
\caption{LLM and human performance on ToM vs factual tasks evaluated using an independent samples test of proportions}
\centering
\begin{tabular}{ccccccc}
\toprule
\multicolumn{1}{c}{\textbf{}} & \multicolumn{1}{c}{\textbf{Task type}} & \multicolumn{1}{c}{\textbf{Trials}} & \multicolumn{1}{c}{\textbf{Successes}} & \multicolumn{1}{c}{\textbf{Mean correct}} & \multicolumn{1}{c}{\textbf{Standard error}} & \multicolumn{1}{c}{\textbf{}} \\
\midrule
LaMDA & ToM & 280 & 140 & 50.0 & .030 & $Z = .000$ \\
 & factual & 280 & 140 & 50.0 & .030 & $p = .5, N = 560$\\
 \cmidrule(r){1-7}
PaLM & ToM & 280 & 166 & 59.3 & .029 & $Z = -.086$\\
 & factual & 280 & 167 & 59.6 & .029 & $p = .466, N = 560$ \\
  \cmidrule(r){1-7}
Flan-PaLM & ToM & 280 & 236 & 84.3 & .022 & $Z = -.3502$\\
 & factual & 280 & 262 & 93.8 & .015 & $\textit{\textbf{p = <.001}}, N = 560$ \\
  \cmidrule(r){1-7}
GPT-3.5 & ToM & 280 & 147 & 52.5 & .030 & $Z = -2.480$\\
 & factual & 280 & 176 & 62.9 & .029 & $\textit{\textbf{p = .007}}, N = 560$ \\
  \cmidrule(r){1-7}
GPT-4 & ToM & 280 & 248 & 88.6 & .019 & $Z = -2.415$\\
 & factual & 280 & 264 & 94.3 & .014 & $\textit{\textbf{p = .008}}, N = 560$ \\
  \cmidrule(r){1-7}
Humans & ToM & 280 & 253 & 90.4 & .018 & $Z = -3.539$\\
 & factual & 280 & 273 & 97.5 & .009 & $\textit{\textbf{p = <.001}}, N = 560$ \\
\bottomrule
\end{tabular}
\label{tab:example}
\end{table}

\end{document}